\documentclass[runningheads]{llncs}

\usepackage{accv}

\usepackage{eso-pic}

\usepackage{accvabbrv}

\usepackage[T1]{fontenc} 
\usepackage{graphicx}
\usepackage{booktabs}
\usepackage{subcaption} 
\usepackage[export]{adjustbox} 
\usepackage{multirow}

\usepackage[accsupp]{axessibility}  

\usepackage{hyperref}
\usepackage{color}

\usepackage{orcidlink}

\begin{document}

\AddToShipoutPictureBG*{%
  \AtPageLowerLeft{%
    \raisebox{1cm}{%
      \makebox[\paperwidth][c]{%
        \parbox{0.8\paperwidth}{\footnotesize\centering
          This is a post-peer-review, pre-copyedit version of an article
          published in ACCV 2024. The final authenticated version
          is available online at \url{https://doi.org/10.1007/978-981-96-0911-6_24}.
        }%
      }%
    }%
  }%
}

\title{Reference-Based Face Super-Resolution Using the Spatial Transformer} 

\titlerunning{FSRST}

\author{Varun Ramesh Jois\orcidlink{0009-0004-2615-6406},
Antonella DiLillo,
James Storer}

\authorrunning{Jois et al.}

\institute{Brandeis University, Waltham MA 02453, USA
\email{\{vjois,dilant,storer\}@brandeis.edu}}

\maketitle

\begin{abstract}
  Face super-resolution is the task of increasing the resolution of an image containing a face thereby adding finer detail. It is a ubiquitous task in many computer vision applications and quite often the user isn’t even aware that it is being performed. However, doing it with high fidelity is challenging as it is an ill-posed problem. In this paper we present a reference-based solution for face super-resolution that uses higher resolution reference images to aid in the task. We show an alignment module based on the spatial transformer that is considerably more stable than the popular deformable convolutions. We also show an aggregation function that can take good quality information from the reference images when available or suppress the function when such information is unavailable. Finally, we show that our relatively smaller model can achieve state of the art results on multiple datasets. The source code is available at \url{https://github.com/varun-jois/FSRST}.
\end{abstract}

\section{Introduction}
\label{sec:intro}
Super-resolution is the task of taking an image and increasing its resolution. For instance, if we have a $100\times100$ pixel image, and we convert it to a $400\times400$ pixel image, what we have done is perform 4$\times$ super-resolution. This has the effect of increasing the finer details in an image leading to a more visually pleasing image. Super-resolution is a fundamental task in low-level computer vision and most image and video applications have some functionality for it. In fact, it is so universal a task that oftentimes, the user isn't even aware that it is being performed in the background. The super-resolution task is challenging, especially when upsampling by a large factor such as 4$\times$ and 8$\times$. The main issue being that it is an ill-posed problem where a single input could potentially map to different outputs. 

To counteract the ill-posedness of super-resolution, many techniques have been suggested to constrain the output of the model. One of these methods is to use one or more high-resolution reference images that are similar in content and texture to the image that is being super-resolved, thereby aiding the task. This is the study known as reference-based super-resolution. Another method to tighten the definition of the task is to put constraints on the data the model is being shown. This happens naturally when we train for a particular type of data such as medical images or satellite images. When we constrain the data to images of faces, we call this face super-resolution. In this paper we perform face super-resolution with the help of high-resolution reference images where the reference images are of the same person. This idea is intriguing for a number of reasons. First, we reduce the difficulty of the problem by constraining the output to textures and shapes found in the reference images and those commonly found in faces. Second, high-resolution images of faces are in abundance whether they be on social media or stored in a user's device making it a practical approach. Third, this idea can directly be applied to compressing video in video conferencing applications that have seen a surge in usage in recent years\cite{SUDUC2022288, Tudor_Cristiana, Kristóf_2020}.  

However, using high-resolution reference images introduces many new challenges: How do we find similar shapes and concepts in the reference \ie the correspondence problem? How do we deal with the discrepancy in resolution? How do we combine/aggregate information from the input and references? If there are multiple references, how do we weight their importance? These are non-trivial problems, some of which are their own field of study.

One popular approach for handling the correspondence issue is by aligning the reference and input images. This is commonly seen in video super-resolution models\cite{Tian_Yapeng_tdan, Wang_Xintao_EDVR, Chan_Kelvin_basicvsr, Chan_Kelvin_basicvsrpp, Chan_Kelvin_2021}. If the input image and the reference image are very similar and can be brought to the same resolution, then alignment would provide the means for \textit{implicit} correspondence matching. Unlike key-point matching, where individual points from the two images are matched \textit{explicitly}, with implicit matching the points are matched as a consequence of being aligned. This could only work if the two images are very similar such as consecutive frames in a video or face images of the same identity. However, a big advantage of this method is that it is fast. The most popular method for alignment in super-resolution is deformable alignment\cite{hime, Tian_Yapeng_tdan, Wang_Xintao_EDVR, Chan_Kelvin_basicvsrpp, Chan_Kelvin_2021} that makes use of the deformable convolution\cite{Dai_Jifeng_dcn, Zhu_Xizhou_dcnv2}. But as pointed out in\cite{hime, Chan_Kelvin_2021, Chan_Kelvin_basicvsrpp}, the deformable convolution is hard to train for alignment due to instability issues and frequently results in training collapse. To address this shortcoming, we develop a novel alignment module that is based on the spatial transformer\cite{jaderberg_spatial} that produces good alignment results and is free from instability. To the best of our knowledge, our paper is the first to show how the spatial transformer can be used for aligning a reference image in super-resolution tasks.

For the issue of information aggregation, we devise a new aggregation scheme that is fast and prioritizes references that are more similar to the input low-resolution image. It does this by weighting the references based on the $l^2$-distance from the input with those closer in distance getting a larger weight. Our aggregation module is flexible with the ability to take one or more reference images. It is designed to not only give a larger weight to more similar references but to also ignore all the references when none of them are a good match. This is helpful when similar reference images are unavailable and so the model can simply perform single image super-resolution (SISR) without being hindered by dissimilar references. 

Our contributions are the following: 
\begin{enumerate}
    \item We present a new method for aligning images that is considerably more stable than the deformable convolution and can be used in other tasks such as video super-resolution.
    \item We show a new aggregation technique based on the $l^2$-distance that can not only prioritize the more similar reference image but can also ignore all references when none are similar.
    \item We introduce the \textbf{F}ace \textbf{S}uper-\textbf{R}esolution using the \textbf{S}patial \textbf{T}ransformer (FSRST) model; a novel lightweight model for reference-based face super-resolution that outperforms state of the art models on multiple datasets.
\end{enumerate}

\section{Related Work}
\subsection{Reference-Based Super-Resolution and Face Super-Resolution}
In recent years there has been a tremendous amount of interest shown in reference based methods for super-resolution\cite{srntt, hime, ttsr, mrefsr, c2, gwainet, asffnet}. One of the earliest models for reference-based super-resolution in the deep-learning era was SRNTT\cite{srntt} that was inspired by the style transfer problem. The Texture Transformer Network for Image Super-Resolution (TTSR)\cite{ttsr} was inspired by the transformer architecture\cite{attention} and consists of a hard and soft attention module to perform super-resolution in a cross-scale manner. The $C^2$-Matching network\cite{c2} was designed to handle the transformation gap as well as the resolution gap between the low-resolution input and the high-resolution reference images. For the transformation gap they solve the correspondence problem with clever data augmentation and for the resolution gap they implement a type of knowledge distillation. The Multi-Reference Super-Resolution model (MRefSR)\cite{mrefsr} was inspired by $C^2$-Matching but is designed to use multiple reference images. They perform feature fusion using their Multi-Reference Attention Module and do feature selection with the help of the Spatial Aware Filtering Module. The Headshot Image Super-Resolution with Multiple Exemplars network (HIME)\cite{hime} uses the deformable convolution for alignment and does aggregation based on their content-conditioned feature aggregation scheme.

\subsection{Problems Aligning with Deformable Convolutions}
Deformable convolutions \cite{Dai_Jifeng_dcn} were designed to alter the sampling locations of the convolution kernel to provide greater transformation ability to the plain convolution kernel. By not fixing the sampling locations, the model is able to learn offsets to the sampling locations allowing the kernel to accommodate different shapes within the input image. In DCNv2 \cite{Zhu_Xizhou_dcnv2} a modulation mask was added to further enhance the modeling capabilities. 

The first uses of deformable convolutions for alignment were for the video super-resolution task \cite{Tian_Yapeng_tdan, Wang_Xintao_EDVR}. In TDAN\cite{Tian_Yapeng_tdan}, deformable convolutions were used for the temporal alignment of frames. In EDVR\cite{Wang_Xintao_EDVR} a Pyramid Cascading Deformable alignment was proposed to align at multiple scales. BasicVSR$++$\cite{Chan_Kelvin_basicvsrpp} uses optical flow to guide the deformable alignment of frames. HIME\cite{hime} was the first model to use deformable convolutions for aligning reference images in the reference-based image super-resolution task.

However, as pointed out in \cite{Chan_Kelvin_2021, hime, Chan_Kelvin_basicvsrpp}, deformable convolutions used for alignment can be difficult to train. Training instability can cause offsets to overflow leading to model degeneration. State of the art reference and video super-resolution models that use deformable alignment need to artificially constrain the offsets to small deviations from the optical flow \cite{Chan_Kelvin_2021, Chan_Kelvin_basicvsrpp, hime} in order to counter the training instability. This necessitates having an additional model to estimate the optical flow therefore increasing the time and space requirements.

From our experience working with deformable convolutions for alignment, training was extremely unstable. Without any guidance for the offsets, training loss could wildly fluctuate and the training can unexpectedly fail even after several hours. The learning rate for the deformable convolution parameters had to be carefully selected along with the right seed in order to train the model.

\section{FSRST Framework}
Our model was inspired by HIME\cite{hime}, another model for reference-based face super-resolution. While reproducing HIME we noticed a number of issues which led us to the architecture of our model. First, we noticed that the deformable alignment was very unstable to train. This was mentioned by the authors but nevertheless, gave us an avenue to explore. Our alignment module addresses this using the spatial transformer. Second, the content-conditioned feature aggregation module in HIME was causing a division by zero in a few instances leading to a degenerate model. Our solution for weighted aggregation was designed to overcome this issue.

\begin{figure*}[tb]
    \centering
    \includegraphics[width=\linewidth]{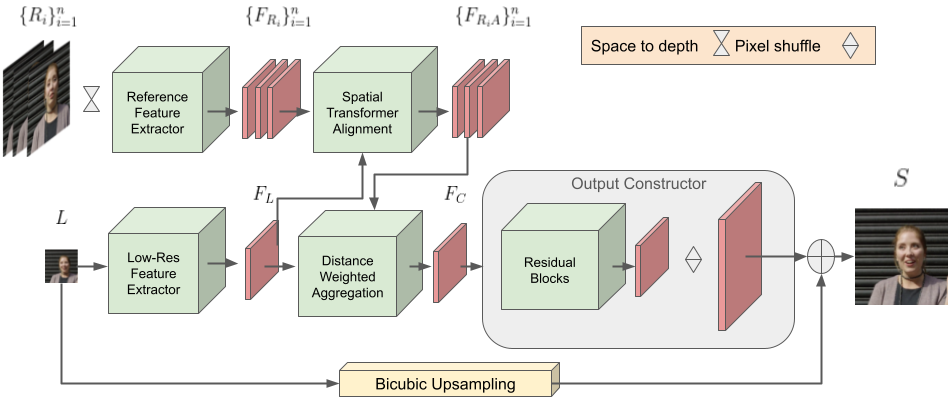}
    \caption{The Face Super Resolution using the Spatial Transformer model (FSRST). Our model takes a low-resolution image $L$ of a person, and $n$ high-resolution reference images $\{R_i\}_{i=1}^{n}$ of the same person and produces super-resolved image $S$.}
    \label{fig:framework}
\end{figure*}

Our model, visualized in \cref{fig:framework}, comprises four parts; the feature extractors, the spatial transformer alignment (STA) module, distance-based weighted aggregation (DWA) and the output constructor. The model takes as input a low resolution image $L$ and a set of high resolution reference images $R = \{R_1, R_2, \ldots R_n\}$ where $n$ is the number of reference images, to produce the super-resolved output $S$. We now discuss the individual parts of the model and how the data flows through the model.

\subsection{Feature Extractor}
Our feature extractor is composed of 5 residual blocks\cite{he_kaiming_resnet} that extract features from the provided inputs. We use different feature extractors for input $L$ and references $R$. The feature extractor for $L$ produces the feature map $F_L$. For the references, we first convert the images from RGB to grayscale since we are mostly concerned with shape information and to improve efficiency. We then perform a space to depth operation \cite{shi_wenzhe_shuffle} to obtain the same spatial resolution as $L$ without losing any information. Finally, we pass the references to the feature extractor to obtain the feature maps $F_R = \{F_{R_1}, F_{R_2}, \ldots F_{R_n}\}$.

\subsection{Spatial Transformer Alignment}
\label{sec:sta}
In order to obtain the most information from the references, it is crucial we align the low-resolution features $F_L$ and the reference features $F_R$. In order to do that we introduce a new alignment module based on the spatial transformer\cite{jaderberg_spatial}. First let us review the spatial transformer.

The spatial transformer consists of three parts, a localisation network, a grid generator and a sampler. The localisation network $f_l$ is a small neural network that takes the input feature map $F \in \mathbb{R}^{H \times
W \times C}$ and produces a transformation matrix $\theta$. 
\begin{equation}
    \theta = f_l(F)
\end{equation}
Depending on the transformation being performed, the dimensions of $\theta$ will vary; for instance if a linear transformation is being performed $\theta$ will be a $2\times2$ matrix, or if an affine transformation is desired $\theta$ will be a $2\times3$ matrix, etc. The grid generator $f_g$ takes as input the transformation matrix $\theta$ and produces the sampling grid $G$. 
\begin{equation}
    G = f_g(\theta)
\end{equation}
In the two dimensional space such as images, $G$ is of shape $H \times W \times 2$ where the 2 components of the last dimension are the horizontal and vertical sampling locations. In the case of multi-channel feature maps, the same sampling grid is used for all channels. Finally, the sampler takes the input feature map $F$ and the sampling grid $G$ to warp/transform $F$ into $F_{T}$ using the input values from $F$ and the pixel locations as determined by $G$:
\begin{equation}
    F_T = warp(F, \ G)
\end{equation}
For points $G$ that are in between pixel locations an interpolation method is used and for out-of-bound grid locations (\eg when a translation is performed) a padding mode is chosen. In our experiments we used bi-linear interpolation and reflection padding. It is worth noting that the grid generator and sampler are parameter free and fully differentiable.

\begin{figure*}[tb]
    \centering
    \includegraphics[width=\linewidth]{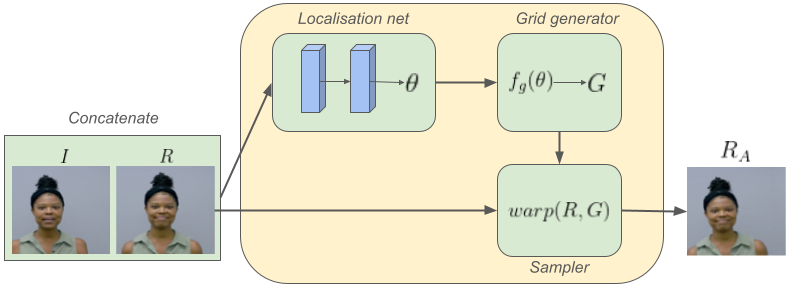}
    \caption{Spatial Transformer Alignment (STA). Here we show how we can align a reference image $R$ to an input image $I$ to produce aligned image $R_{A}$. In our model FSRST, we perform the alignment in feature space.}
    \label{fig:alignment}
\end{figure*}

We use the spatial transformer in our model to perform an affine transformation on each reference feature $\{F_{R_i}\}_{i=1}^{n}$ to align them with the low-resolution features $F_L$. \Cref{fig:alignment} shows our design. We do this by first concatenating $F_L$ and $F_{R_i}$ along the channel dimension. We then pass it to the localisation network $f_l$ that comprises of two 32-filter 5$\times$5 convolution layers with a 2$\times$2 max-pooling layer in between followed by a 32 unit and 6 unit fully-connected layer producing a vector output. The output is then reshaped to form the transformation matrix $\theta$: 
\begin{equation}
    \theta = f_l(Concatenate(F_L, F_{R_i}))
\end{equation}
Since we are performing an affine transformation $\theta$ is a $2\times3$ matrix. We pass $\theta$ to the grid generator $f_g$ to create our sampling grid $G$. Finally, the sampler uses $G$ to warp our reference image $F_{R_i}$ and produce the aligned reference $F_{R_iA}$:
\begin{equation}
    F_{R_iA} = warp(F_{R_i}, \ G)
\end{equation}
We align each of the reference features this way to produce the aligned features $F_{RA} = \{F_{R_1A}, F_{R_2A}, \ldots F_{R_nA}\}$.

\subsection{Distance-Based Weighted Aggregation}
Now that we have the features for the low-resolution input $F_L \in \mathbb{R}^{H \times W \times C}$ and the aligned reference features $\{F_{R_iA}\}_{i=1}^{n}$ where $F_{R_iA} \in \mathbb{R}^{H \times W \times C}$  we need to find a way to combine or aggregate them. Ideally, for each spatial location in the input feature map ($H\times W$) we want to take the reference features that are most similar to the input at that location and ignore the reference features that are least similar. We do this by combining the $l^2$-distance and the softmax function.

First, for each aligned reference feature $\{F_{R_iA}\}_{i=1}^{n}$ we do a component-wise subtraction with $F_L$ and calculate the $l^2$-distance $D_i$ for each pixel location:

\begin{equation}
    R_i = F_L - F_{R_iA}
\end{equation}

\begin{equation}
    D_i = \sqrt{\sum_{k=1}^{C}r_{ijk}^2}, \quad D_i \in \mathbb{R}^{H \times W}
\end{equation}
We then concatenate each of the spatial location distances along the channel dimension and compute the softmax of the negative along the channel dimension. This gives us our weights for aggregation $WT$.

\begin{equation}
    D = Concatenate(D_1, D_2, \ldots D_n), \quad D \in \mathbb{R}^{H \times W \times n}
    \label{eq:clamp}
\end{equation}

\begin{equation}
    WT = Softmax(-D) = \frac{e^{-d_{ijk}}}{\sum_{k=1}^{n}e^{-d_{ijk}}}, \quad WT \in \mathbb{R}^{H \times W \times n}
    \label{eq:eps}
\end{equation}
The reason we take the softmax of the negative is because we are calculating the softmax for $l^2$-distances and we want to give greater weight to smaller distances. Each channel $i$ in $WT$ is the corresponding weight for aligned reference feature $F_{R_iA}$.  We get the aggregated references $F_{agg}$, by multiplying the weights component-wise to the references and adding. 

\begin{equation}
    F_{agg} = \sum_{i=1}^{n} WT_{i} \ . \ F_{R_iA}, \quad F_{agg} \in \mathbb{R}^{H \times W \times C}
\end{equation}
where $WT_{i}$ is the $i$'th channel of $WT$.

To further improve the performance of our aggregation, we clamp the values of the $l^2$-distances in \cref{eq:clamp} to the range $[0, 10^2]$. We also add an $\epsilon = 10^{-9}$ to the denominator of \cref{eq:eps}. These two changes provide additional stability to the training and also creates the effect of ignoring all the references when the $l^2$-distance is large for all of them (by making their weights zero). Finally we combine the low-resolution input feature map and the aggregated features and pass it on to the Output Constructor:
\begin{equation}
    F_C = F_L + F_{agg}
\end{equation}

\subsection{Output Constructor}
The final part of our model takes the combined features $F_C$ produced by the aggregation module and constructs the final output. First $F_C$ is passed through a sequence of 20 residual blocks before a channel-to-space transformation is performed using the sub-pixel convolution\cite{shi_wenzhe_shuffle}. This produces a super-resolved version of the input. But from our experiments and from those in \cite{hime}, the output constructor underperforms when it directly tries to predict the output. Instead, we first perform bicubic upsampling on the low-resolution input $L$ and make the output constructor $f_c$ predict a residual which gets added to make the final super-resolved output $S$:
\begin{equation}
    S = bicubic(L) + f_c(F_C)
\end{equation}

\section{Experiments}
\label{sec:exp}
\subsection{Datasets}
We work with the the DeepFakeDetection\cite{roessler2019faceforensicspp}, the CelebAMask-HQ\cite{Lee_2020_CVPR}  and VoxCeleb2\cite{chung18b} datasets for our experiments which are all publicly available. We now explain how the training and validation sets were constructed for each of these datasets. For our super-resolution experiments, we used three reference images so each training sample consisted of the ground-truth image, the low-resolution version of the ground-truth \ie the input image and three reference images of the same person. 

\subsubsection{DeepFakeDetection (DFD).} 
\label{sec:dfd}
The DeepFakeDetection dataset\cite{roessler2019faceforensicspp} is comprised of 363 professionally taken videos of 28 paid actors in various settings. Similar to \cite{Agnolucci} the first 22 identities were used for training and the remaining 6 were used for validation. Among all the classes of videos, we only considered the classes containing the title \textit{"outside talking still laughing"}, \textit{"podium speech happy"} or \textit{"talking against wall"} as these were the closest to real-life video calls. We then manually took a $512\times512$ pixel crop that centred the head in each video. The frames 0, 48 and 96 were chosen as reference images and 20 frames starting from frame 192 with an interval of 20 frames were selected as the ground-truth frames to our model. We finally downsampled all the images using bicubic downsampling to a resolution of $128\times128$ pixels. To produce the low-resolution input images, we further downsampled the ground-truth images to $32\times32$ pixels. In total we had 1300 samples for training and 340 samples for testing. 

\subsubsection{CelebAMask-HQ.}
\label{sec:celeba}
In addition to the DFD dataset, we use the CelebAMask-HQ dataset \cite{Lee_2020_CVPR} for training and evaluation. It is based off of the CelebA-HQ dataset \cite{karras2018progressive} which in turn was created from the CelebA dataset \cite{Liu_CelebA}. The dataset is comprised of 30,000 $1024\times1024$ pixel images of various celebrities. We first obtained identity information from \cite{Liu_CelebA} and proceeded to create our dataset in a similar fashion to \cite{hime}. We first filtered out identities that had fewer than 4 images leaving us with 2887 identities, each being one image sample. From this we randomly selected 2600 samples for training and 287 for evaluation. We finally formed our data by bicubicly downsampling the images to $128\times128$ pixels and further downsampling the ground-truth to $32\times32$ pixels to form our low-resolution input. 

\subsubsection{VoxCeleb2.}
To further test our model, we built an additional evaluation set from the VoxCeleb2 dataset \cite{chung18b}. The VoxCeleb2 test set consists of 118 identities and 4,911 videos which are split into 36,237 utterances or short snippets. To produce our image dataset, for each identity, we randomly selected 4 videos to be our ground-truth plus 3 references and then extracted the first frame for each video. Again, we use bicubic downsampling to produce $128\times128$ pixel images and we further downsampled the ground-truth images to $32\times32$ pixels to produce our low-resolution input.

\subsection{Super-Resolution Experiments}
To test our model on the super-resolution task we trained two models; one on the DeepFakeDetection (DFD) dataset and another on the CelebAMask-HQ dataset. The DFD dataset gives us a good approximation of performance in a video conferencing setting as the reference images are very similar to the images we are trying to super-resolve whereas the CelebAMask-HQ dataset gives us a good approximation of performance on a general purpose face dataset where we have multiple images of the same identity but of varying similarity. To further evaluate our model, we use the model trained on the CelebAMask-HQ dataset and test it on the VoxCeleb2 dataset.

We compare our model to 4 recent state of the art models that were each trained on the DeepFakeDetection and CelebAMask-HQ datasets. Each of these models use reference images to perform super-resolution with a varying number of reference images. The Texture Transformer Network for Image Super-Resolution (TTSR)\cite{ttsr} and the $C^2$-Matching network\cite{c2} use one reference image. The Multi-Reference Super-Resolution model (MRefSR)\cite{mrefsr} and the Headshot Image Super-Resolution with Multiple Exemplars network (HIME)\cite{hime} can take an arbitrary number of reference images. To match our training conditions we give MRefSR and HIME $n=3$ reference images. For TTSR, MRefSR and $C^2$-Matching we used the code that was publicly released by the respective authors to train the models. For HIME, no code was publicly released so we had to write code for it from scratch. We recreated the HIME-small model which purely relies on the deformable convolution for alignment but we had to give it more parameters for better performance.

For all the models tested, we optimized for recreation loss only and did not perform adversarial training. While adversarial training would have produced more visually pleasing results, it is still challenging to measure perceptual quality and all the available perceptual quality metrics have their pros and cons. Also, obtaining mean opinion scores from humans is an expensive process so we decided to forego training for perceptual quality and only optimize for pixel-wise recreation loss. In our experiments, we used the $L1$-loss: 
\begin{equation}
    \mathcal{L} = \frac{1}{HWC}\left \lVert G - S \right \rVert_1
\end{equation}
where $G$ is the ground-truth and $S$ is the super-resolved image. $HWC$ is the height, width and number of channels in $G$. In all experiments, we performed $4\times$ super-resolution on a low-resolution image of size $32\times32$ pixels.

\begin{table}[tb]
  \caption{Performance scores for Super-resolution. Best scores are shown in \textbf{bold}.
  }
  \label{tab:sr_scores}
  \centering
  \begin{tabular}{c|c|c|c|c|c}
    \toprule
    Model & Training set & Testing set & PSNR (↑) & SSIM (↑) & Parameters (M) \\
    \midrule
    TTSR & DFD & DFD & 31.2112 & 0.9238 & 6.73 \\
    MRefSR & DFD & DFD & 31.626 & 0.9252 & 24.00 \\
    $C^2$-Matching & DFD & DFD & 32.0086 &  \textbf{0.9339} & 9.98 \\
    HIME & DFD & DFD & 31.0227 & 0.9178 & 2.16 \\
    Ours (small) & DFD & DFD & 31.7224 & 0.9261 &  \textbf{0.87} \\
    Ours & DFD & DFD &  \textbf{32.0838} & 0.9308 & 2.55 \\
    \hline
    TTSR & CelebA & CelebA & 28.8223 & 0.8658 & 6.73 \\
    MRefSR & CelebA & CelebA & 27.9561 & 0.8419 & 24.00 \\
    C2-Matching & CelebA & CelebA & 28.3862 & 0.8571 & 9.98 \\
    HIME & CelebA & CelebA & 29.1198 & 0.8709 & 2.16 \\
    Ours (small) & CelebA & CelebA & 29.0562 & 0.8693 & \textbf{0.87} \\
    Ours & CelebA & CelebA & \textbf{29.2842} & \textbf{0.8739} & 2.55 \\
    \hline
    TTSR & CelebA & Vox2 & 31.4478 & 0.917 & 6.73 \\
    MRefSR & CelebA & Vox2 & 30.5125 & 0.8974 & 24.00 \\
    C2-Matching & CelebA & Vox2 & 30.8582 & 0.9092 & 9.98 \\
    HIME & CelebA & Vox2 & 31.6232 & 0.9181 & 2.16 \\
    Ours (small) & CelebA & Vox2 & 31.6064 & 0.9174 & \textbf{0.87} \\
    Ours & CelebA & Vox2 & \textbf{31.7547} & \textbf{0.9201} & 2.55 \\
  \bottomrule
  \end{tabular}
\end{table}

\subsubsection{Quantitative Results.}
\Cref{tab:sr_scores} shows the performance of the various models on the super-resolution task. It is divided into three sections based on the dataset used for training and the dataset used for testing. We evaluate the models based on PSNR and SSIM \cite{ssim}. We do not report scores for a perceptual metric because we can either optimize for low distortion or high perceptual quality but not both\cite{per_dist}. We chose to optimize for low distortion to preserve identity information. The best scores are marked in bold. For the DFD dataset, our model was able to outperform the second best model $C^2$-Matching by nearly 0.08 dB and the other models by an average margin of nearly 0.8 dB on the PSNR metric. On the CelebAMask-HQ dataset (denoted as CelebA in the table to reduce space), our model outperformed the next best model HIME by nearly 0.17 dB and the other models by an average margin of nearly 0.9 dB. On the VoxCeleb2 test set (denoted as Vox2 in the table), our model outperformed the second best model by over 0.13 dB and the other models by over 0.8 dB on average.

\Cref{tab:sr_scores} also shows the size of each model in terms of the number of parameters. This can be found in the last column. Even though our model is considerably smaller than TTSR, MRefSR and $C^2$-Matching it outperforms all of them on PSNR for all three test sets. This shows that with even a light-weight model, good performance can be achieved when alignment and aggregation are performed well. 

In our experiments we also constructed a small version of our model which had the exact same architecture as the larger model except with fewer channels in the residual blocks. In the table this model is referred to as Ours (small). Even though this model had only a fraction of the number of parameters as compared to the other models, it was still able to outperform three of the four models. This light-weight model would be well suited for mobile devices where there are space and processing constraints.

\begin{figure}[tb]
    \centering
    \captionsetup[subfigure]{labelformat=empty}
    \begin{subfigure}{0.15\linewidth}
        \includegraphics[width=\linewidth, valign=c]{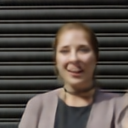}
    \end{subfigure}
    \begin{subfigure}{0.15\linewidth}
        \includegraphics[width=\linewidth, valign=c]{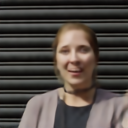}
    \end{subfigure}
    \begin{subfigure}{0.15\linewidth}
        \includegraphics[width=\linewidth, valign=c]{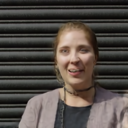}
    \end{subfigure}
    \begin{subfigure}{0.15\linewidth}
        \includegraphics[width=\linewidth, valign=c]{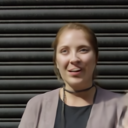}
    \end{subfigure}
    \begin{subfigure}{0.15\linewidth}
        \includegraphics[width=\linewidth, valign=c]{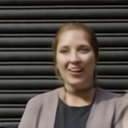}
    \end{subfigure}
    \begin{subfigure}{0.15\linewidth}
        \includegraphics[width=\linewidth, valign=c]{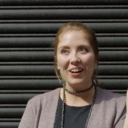}
    \end{subfigure}

    \begin{subfigure}{0.15\linewidth}
        \includegraphics[width=\linewidth, valign=c]{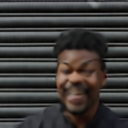}
    \end{subfigure}
    \begin{subfigure}{0.15\linewidth}
        \includegraphics[width=\linewidth, valign=c]{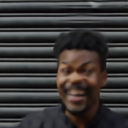}
    \end{subfigure}
    \begin{subfigure}{0.15\linewidth}
        \includegraphics[width=\linewidth, valign=c]{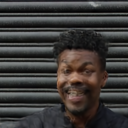}
    \end{subfigure}
    \begin{subfigure}{0.15\linewidth}
        \includegraphics[width=\linewidth, valign=c]{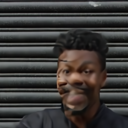}
    \end{subfigure}
    \begin{subfigure}{0.15\linewidth}
        \includegraphics[width=\linewidth, valign=c]{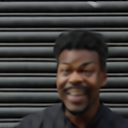}
    \end{subfigure}
    \begin{subfigure}{0.15\linewidth}
        \includegraphics[width=\linewidth, valign=c]{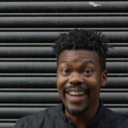}
    \end{subfigure}

    \begin{subfigure}{0.15\linewidth}
        \includegraphics[width=\linewidth, valign=c]{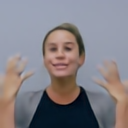}
        \subcaption{HIME}
    \end{subfigure}
    \begin{subfigure}{0.15\linewidth}
        \includegraphics[width=\linewidth, valign=c]{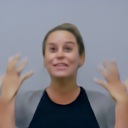}
        \subcaption{TTSR}
    \end{subfigure}
    \begin{subfigure}{0.15\linewidth}
        \includegraphics[width=\linewidth, valign=c]{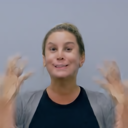}
        \subcaption{MRefSR}
    \end{subfigure}
    \begin{subfigure}{0.15\linewidth}
        \includegraphics[width=\linewidth, valign=c]{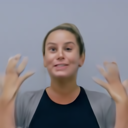}
        \subcaption{C2-Matching}
    \end{subfigure}
    \begin{subfigure}{0.15\linewidth}
        \includegraphics[width=\linewidth, valign=c]{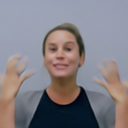}
        \subcaption{Ours}
    \end{subfigure}
    \begin{subfigure}{0.15\linewidth}
        \includegraphics[width=\linewidth, valign=c]{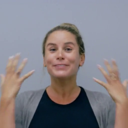}
        \subcaption{HR}
    \end{subfigure}
    \begin{subfigure}{0.15\linewidth}
        \includegraphics[width=\linewidth, valign=c]{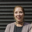}
    \end{subfigure}
    \begin{subfigure}{0.15\linewidth}
        \includegraphics[width=\linewidth, valign=c]{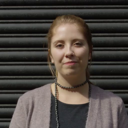}
    \end{subfigure}
    \begin{subfigure}{0.15\linewidth}
        \includegraphics[width=\linewidth, valign=c]{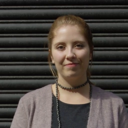}
    \end{subfigure}
    \begin{subfigure}{0.15\linewidth}
        \includegraphics[width=\linewidth, valign=c]{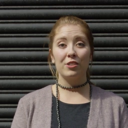}
    \end{subfigure}

    \begin{subfigure}{0.15\linewidth}
        \includegraphics[width=\linewidth, valign=c]{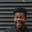}
    \end{subfigure}
    \begin{subfigure}{0.15\linewidth}
        \includegraphics[width=\linewidth, valign=c]{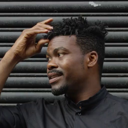}
    \end{subfigure}
    \begin{subfigure}{0.15\linewidth}
        \includegraphics[width=\linewidth, valign=c]{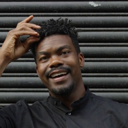}
    \end{subfigure}
    \begin{subfigure}{0.15\linewidth}
        \includegraphics[width=\linewidth, valign=c]{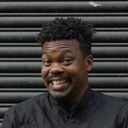}
    \end{subfigure}

    \begin{subfigure}{0.15\linewidth}
        \includegraphics[width=\linewidth, valign=c]{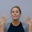}
        \subcaption{LR}
    \end{subfigure}
    \begin{subfigure}{0.15\linewidth}
        \includegraphics[width=\linewidth, valign=c]{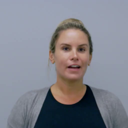}
        \subcaption{Ref-1}
    \end{subfigure}
    \begin{subfigure}{0.15\linewidth}
        \includegraphics[width=\linewidth, valign=c]{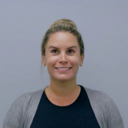}
        \subcaption{Ref-2}
    \end{subfigure}
    \begin{subfigure}{0.15\linewidth}
        \includegraphics[width=\linewidth, valign=c]{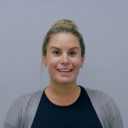}
        \subcaption{Ref-3}
    \end{subfigure}
  
    \caption{Super-resolution results from the DFD test set. Here we super-resolved the low-resolution image \textit{LR} with the support of high-resolution references.} 
    \label{fig:sr1}
\end{figure}


\subsubsection{Qualitative Results.}
\Cref{fig:sr1} shows a sample of the outputs for the super-resolution task. These are examples from the test set of the DFD dataset. The figure is divided into two parts where the first part contains the output from the models and the second part contains the inputs that were provided. For TTSR and $C^2$-Matching that only took one reference image the first reference image was used. In all the cases, our model does a faithful job recreating the ground truth from the low-resolution input. In the first row, HIME and TTSR have difficulties producing the mouth whereas MRefSR and $C^2$-Matching incorrectly produce the eyes giving an uncanny valley effect. In the second row, HIME and TTSR produce very blurred eyes, MRefSR introduces speckles and $C^2$-Matching completely deforms the face. In the third row, HIME incorrectly produces the mouth whereas TTSR, MRefSR and $C^2$-Matching changes the shape of the eyes again causing an uncanny valley effect. 

\begin{table}[tb]
  \caption{Performance for Super-resolution by number of reference images.}
  \label{tab:ref_num}
  \centering
  \begin{tabular}{c|c|c}
    \toprule
    Reference Images & PSNR (↑) & SSIM (↑) \\
    \midrule
    0 & 31.7495 & 0.93 \\
    1 & 31.9437 & 0.9293 \\
    3 &  \textbf{32.0838} &  \textbf{0.9308} \\
    \bottomrule
  \end{tabular}
\end{table}

\subsection{Ablation Studies}
To test the impact of the number of reference images on super-resolution performance we trained a model taking no reference images and a model taking one reference image. For the model taking no references, we created a model based on the ResNet \cite{he_kaiming_resnet} architecture. For the model taking one reference, we made minimum modifications to our model to take one reference instead of three. To ensure fairness, all models had approximately the same number of parameters. All models were trained on the DFD dataset. 

From \cref{tab:ref_num} we can see the result of this experiment. The model that had three reference images to work with outperformed the model with only one reference by 0.14 dB and outperformed the model taking no references by nearly 0.34 dB. This experiment shows that having high-resolution reference images improves performance and our model is able to perform better when multiple references are available.

\subsection{Alignment Experiments}
To test whether our Spatial Transformer Alignment module (\cref{sec:sta}) could be used for aligning two images, we performed two experiments. For both experiments, the goal was to align the faces. Here we performed \textit{explicit} alignment directly on the inputs as opposed to \textit{implicit} alignment in feature space as seen in our super-resolution model. We used 128$\times$128 pixel images for these experiments and made minimal changes in architecture to accommodate the larger images. 

\begin{figure}[tb]
    \centering
    \captionsetup[subfigure]{justification=raggedright}
    \begin{subfigure}{0.15\linewidth}
      \caption*{Input}
    \end{subfigure}\hfil
    \begin{subfigure}{0.15\linewidth}
        \includegraphics[width=\linewidth, valign=c]{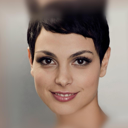}
    \end{subfigure}\hfil
    \begin{subfigure}{0.15\linewidth}
        \includegraphics[width=\linewidth, valign=c]{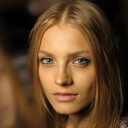}
    \end{subfigure}\hfil
    \begin{subfigure}{0.15\linewidth}
        \includegraphics[width=\linewidth, valign=c]{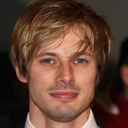}
    \end{subfigure}\hfil
    \begin{subfigure}{0.15\linewidth}
        \includegraphics[width=\linewidth, valign=c]{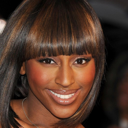}
    \end{subfigure}\hfil
    \begin{subfigure}{0.15\linewidth}
        \includegraphics[width=\linewidth, valign=c]{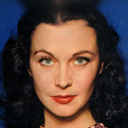}
    \end{subfigure}

    \begin{subfigure}{0.15\linewidth}
      \caption*{Transformed}
    \end{subfigure}\hfil
    \begin{subfigure}{0.15\linewidth}
        \includegraphics[width=\linewidth, valign=c]{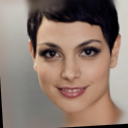}
    \end{subfigure}\hfil
    \begin{subfigure}{0.15\linewidth}
        \includegraphics[width=\linewidth, valign=c]{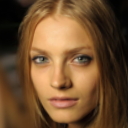}
    \end{subfigure}\hfil
    \begin{subfigure}{0.15\linewidth}
        \includegraphics[width=\linewidth, valign=c]{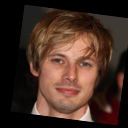}
    \end{subfigure}\hfil
    \begin{subfigure}{0.15\linewidth}
        \includegraphics[width=\linewidth, valign=c]{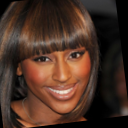}
    \end{subfigure}\hfil
    \begin{subfigure}{0.15\linewidth}
        \includegraphics[width=\linewidth, valign=c]{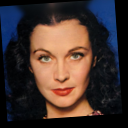}
    \end{subfigure}

    \begin{subfigure}{0.15\linewidth}
      \caption*{Aligned}
    \end{subfigure}\hfil
    \begin{subfigure}{0.15\linewidth}
        \includegraphics[width=\linewidth, valign=c]{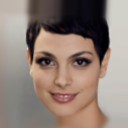}
    \end{subfigure}\hfil
    \begin{subfigure}{0.15\linewidth}
        \includegraphics[width=\linewidth, valign=c]{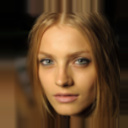}
    \end{subfigure}\hfil
    \begin{subfigure}{0.15\linewidth}
        \includegraphics[width=\linewidth, valign=c]{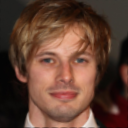}
    \end{subfigure}\hfil
    \begin{subfigure}{0.15\linewidth}
        \includegraphics[width=\linewidth, valign=c]{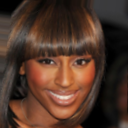}
    \end{subfigure}\hfil
    \begin{subfigure}{0.15\linewidth}
        \includegraphics[width=\linewidth, valign=c]{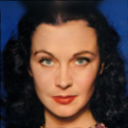}
    \end{subfigure}

    \caption{Alignment results from the first alignment experiment. Here the objective was to align the affine transformed image in the second row to the image in the first row. Results in the third row.}
    \label{fig:align1}
\end{figure}

\subsubsection{Reversing a Known Transformation.}In the first experiment, we tested whether the alignment module could reverse a known affine transformation that was made to an image. Each input image was randomly rotated between $[-10, 10]$ degrees, randomly translated between $[0, 10]$ percent of the image height and width along both directions and randomly scaled between $[0.8, 1.2]$. No shearing was performed. These ranges were chosen to closely simulate the changes seen in a video call. Since the transformation matrix is known, we can easily compute the inverse transformation matrix that would reverse the transformation and align the transformed image with the untransformed one. And since the spatial transformer alignment module calculates the affine matrix $\theta$ that would produce the alignment, we can directly compare it to the known inverse transform. We used the CelebAMask-HQ dataset (\cref{sec:celeba}) for this experiment. We can see the alignment results in \cref{fig:align1}. Row one contains the input image, row two the randomly affine transformed version of the input and row three the result of the alignment module. The objective was to align the images in row two to the images in row one. As we can see here, the alignment module is able to do a near perfect job and this is also reflected in $l^1$ scores where we were able to get a low validation loss of 0.0082.

\subsubsection{Aligning Reference Images.} 
\begin{figure}[tb]
    \centering
    \captionsetup[subfigure]{justification=raggedright}
    \begin{subfigure}{0.15\linewidth}
      \caption*{Input}
    \end{subfigure}\hfil
    \begin{subfigure}{0.15\linewidth}
        \includegraphics[width=\linewidth, valign=c]{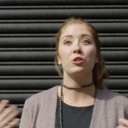}
    \end{subfigure}\hfil
    \begin{subfigure}{0.15\linewidth}
        \includegraphics[width=\linewidth, valign=c]{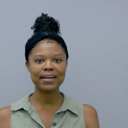}
    \end{subfigure}\hfil
    \begin{subfigure}{0.15\linewidth}
        \includegraphics[width=\linewidth, valign=c]{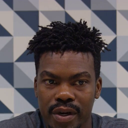}
    \end{subfigure}\hfil
    \begin{subfigure}{0.15\linewidth}
        \includegraphics[width=\linewidth, valign=c]{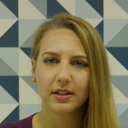}
    \end{subfigure}\hfil
    \begin{subfigure}{0.15\linewidth}
        \includegraphics[width=\linewidth, valign=c]{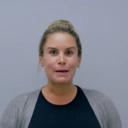}
    \end{subfigure}

    \begin{subfigure}{0.15\linewidth}
      \caption*{Reference}
    \end{subfigure}\hfil
    \begin{subfigure}{0.15\linewidth}
        \includegraphics[width=\linewidth, valign=c]{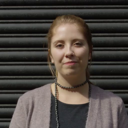}
    \end{subfigure}\hfil
    \begin{subfigure}{0.15\linewidth}
        \includegraphics[width=\linewidth, valign=c]{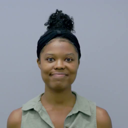}
    \end{subfigure}\hfil
    \begin{subfigure}{0.15\linewidth}
        \includegraphics[width=\linewidth, valign=c]{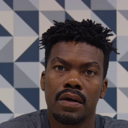}
    \end{subfigure}\hfil
    \begin{subfigure}{0.15\linewidth}
        \includegraphics[width=\linewidth, valign=c]{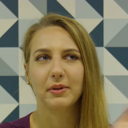}
    \end{subfigure}\hfil
    \begin{subfigure}{0.15\linewidth}
        \includegraphics[width=\linewidth, valign=c]{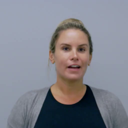}
    \end{subfigure}

    \begin{subfigure}{0.15\linewidth}
      \caption*{Aligned}
    \end{subfigure}\hfil
    \begin{subfigure}{0.15\linewidth}
        \includegraphics[width=\linewidth, valign=c]{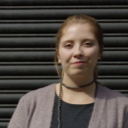}
    \end{subfigure}\hfil
    \begin{subfigure}{0.15\linewidth}
        \includegraphics[width=\linewidth, valign=c]{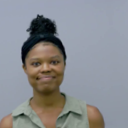}
    \end{subfigure}\hfil
    \begin{subfigure}{0.15\linewidth}
        \includegraphics[width=\linewidth, valign=c]{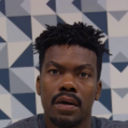}
    \end{subfigure}\hfil
    \begin{subfigure}{0.15\linewidth}
        \includegraphics[width=\linewidth, valign=c]{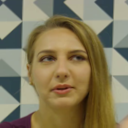}
    \end{subfigure}\hfil
    \begin{subfigure}{0.15\linewidth}
        \includegraphics[width=\linewidth, valign=c]{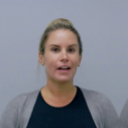}
    \end{subfigure}

    \caption{Alignment results from the second alignment experiment. Here the objective was to align the reference image in the second row to the image in the first row. Results in the third row.}
    \label{fig:align2}
\end{figure}

In our second alignment experiment, we wanted to see if we could align a different but similar image of the same person to an input image. The idea was to closely simulate a video calling experience where the frames are very similar throughout a call. For this experiment we used the DFD dataset (\cref{sec:dfd}). Unlike the previous alignment experiment, the image we were trying to align was different from the input image. So there is no way to know what the best affine transformation would be. So we train by minimizing the pixel-wise $l^1$-loss between the aligned reference image and the input. To focus the optimization on the person rather than the background, we took the loss based on a center-crop of $96\times96$ pixels. The results can be seen in \cref{fig:align2}. The goal was to align the reference images in the second row to the input images in the first row. The alignment results can be seen in the third row. As we can see from the results, the alignment module does a good job performing the affine transformation to align the references. It finds a way to create the greatest overlap between the reference and input images.

\section{Conclusion and Future Work}
In this paper we propose a new model (FSRST) for the task of reference-based face super-resolution. We present a novel alignment module that is based on the spatial transformer that alleviates the instability of deformable convolutions for alignment. Unlike alignment modules based on deformable convolutions, our alignment module doesn't require external guidance in the form of optical flow thereby making it lightweight. We show a novel aggregation methodology that also provides added stability to training. It is capable of extracting useful information from the reference images if found, or simply ignoring the references when unavailable. Finally, our super-resolution model is lightweight but also effective outperforming the other state of the art models on multiple datasets. 

While our alignment module is able to overcome the instability issues presented by deformable convolutions, given the architecture of the spatial transformer, the module is not fully convolutional and can only handle a fixed size input. This is a problem that can be easily overcome by cropping the inputs to a fixed size or by training multiple models for different size inputs but is a shortcoming nonetheless.

For future work we are keen on exploring our model for the task of video super-resolution. We believe our model, with a few changes, can be useful for video compression and is well suited for real-time communication such as video-calls. We also would like to explore ways in which we can make the alignment module fully convolutional.

%
%
\bibliographystyle{splncs04}
\bibliography{main}
\end{document}